\documentclass[journal]{IEEEtran}
\usepackage{graphicx}
\usepackage{cite}
\usepackage{booktabs}
\usepackage{wrapfig}
\usepackage{bigstrut}
\usepackage[normalem]{ulem}
\usepackage{easyReview}
\usepackage{xcolor}
\usepackage{soul}
\usepackage{amssymb}
\usepackage{algorithm}
\usepackage{algorithmic}
\usepackage{stfloats}

\begin{document}
	\title{An Iterative Classification and Semantic Segmentation Network for Old Landslide Detection Using High-Resolution Remote Sensing Images}

	\author{Zili~Lu,
		Yuexing~Peng,~\IEEEmembership{Member,~IEEE,}
		Wei~Li,~\IEEEmembership{Senior~Member,~IEEE,}
		Junchuan~Yu,~Daqing~Ge,~Lingyi~Han,
		and Wei~Xiang,~\IEEEmembership{Senior~Member,~IEEE}
		
		\thanks{Z. Lu and Y. Peng are with the key lab of universal wireless communication, MOE, School of Information and Communication Engineering, Beijing University of Posts and Telecommunications, 100876, Beijing, P.R.China (email: \{luzili0705, yxpeng \}@bupt.edu.cn).}
		
		\thanks{W. Li is with the School of Information and Electronics, Beijing Institute of Technology, 100081, Beijing, P.R China(email: liwei089@ieee.org).}
		
		\thanks{J. Yu, D. Ge and L. Han are with China Aero Geophysical Survey and Remote Sensing Center for Natural Resources, Beijing, 10083, China (email: \{yujunchuan, gedaqing\}@mail.cgs.gov.cn, hanlingyi@bupt.cn).}
		
		\thanks{W. Xiang is with the School of Computing Engineering and Mathematical Sciences, La Trobe University, Melbourne, Australia (email: w.xiang@latrobe.edu.au).}
		
		\thanks{This work was supported in part by National Key Research and Development Program of China under grant 2021YFC3000400 and in part by the High Level Talent Team Project of the New Coast of Qingdao New District under Grant RCTD-JC-2019-06. (Corresponding author: Yuexing Peng.)}
		
		\thanks{Manuscript received **; revised **; accepted **.}
	}
	
	\markboth{IEEE Trans. Geoscience and Remote Sensing, ~Vol.~*, No.~*, *}%
	{Shell \MakeLowercase{\textit{et al.}}: Bare Demo of IEEEtran.cls for IEEE Journals}
	\maketitle
	
	\begin{abstract}
		The geological characteristics of old landslides can provide crucial information for the task of landslide protection. However, detecting old landslides from high-resolution remote sensing images (HRSIs) is of great challenges due to their partially or strongly transformed morphology over a long time and thus the limited difference with their surroundings. Additionally, small-sized  datasets can restrict in-depth learning. 
		To address these challenges, this paper proposes a new iterative classification and semantic segmentation network (ICSSN), which can significantly improve both object-level and pixel-level classification performance by iteratively upgrading the feature extraction module shared by the object classification  and semantic segmentation networks. To improve the detection performance on small-sized datasets, object-level contrastive learning is employed in the object classification network featuring a siamese network to realize global features extraction, and a sub-object-level contrastive learning method is designed in the semantic segmentation network to efficiently extract salient features from boundaries of landslides. An iterative training strategy is also proposed to fuse features in the semantic space, further improving both the object-level and pixel-level classification performances.
		The proposed ICSSN is evaluated on a real-world landslide dataset, and experimental results show that it greatly improves both the classification and segmentation accuracy of old landslides. For the semantic segmentation task, compared to the baseline, the F1 score increases from 0.5054 to 0.5448, the mIoU improves from 0.6405 to 0.6610, the landslide IoU grows from 0.3381 to 0.3743, and the object-level detection accuracy of old landslides surges from 0.55 to 0.90. For the object classification task, the F1 score increases from 0.8846 to 0.9230, and the accuracy score is up from 0.8375 to 0.8875.
	\end{abstract}
	
	\IEEEpeerreviewmaketitle
	
	\section{Introduction}
	Landslides are natural disasters that occur frequently and can cause huge damages, such as casualties, building collapse, and farmland destruction, particularly to human life \cite{shaik2019detection}. Therefore, it is crucial to analyze the statistics of existing landslides for accurate prediction of potential landslides to prevent loss of life and property. Large-scale geological data can efficiently help us understand the characteristics of landslides \cite{ullah2022integrated}. However, traditional methods of obtaining landslide geological data, such as field observation and mapping, are intensively time-consuming and thus cannot be applied in wide-area geologic surveys\cite{DAI200265,PARISE2001697}. Fortunately, with the rapid advancement of remote sensing technology, massive amounts of data have been accumulated and widely used for remote sensing tasks, including land utilization\cite{zhang2022remote}, environment monitoring\cite{li2020review}, agriculture\cite{shanmugapriya2019applications}, forestry\cite{lechner2020applications}, and city planning\cite{bai2022urban}. The same trend has been observed in landslide detection, where remote sensing data is utilized for potential landslide detection and impact assessment \cite{casagli2023landslide}.
	
	For landslide detection using remote sensing data, manual visual interpretation is still the preliminary method, where specialists analyze the geomorphic manifestations of the disaster area and the landslide generation mechanism process then circle the landslides\cite{guzzetti2012landslide}. This kind of time-consuming and labor-intensive method is short of automation, intelligence, and robustness, as it heavily relies on the expertise of specialists\cite{man1landsilde,man2landsilde}.
	
	To improve the efficiency of landslide detection, a maltitude of machine learning-based methods have been proposed. In\cite{keyport2018comparative,7855696,li2016landslide}, methods are proposed to classify each pixel directly based on the spectral information of a high-resolution image. Alternatively, in  \cite{blaschke2010object,martha2010characterising,lahousse2011landslide,holbling2012semi,pawluszek2019multi}, objects with similar features are classified using spectral, spatial, hierarchical, and other features to extract landslide information. Although these machine learning methods initially achieve a wide range of automatic landslide detection, which greatly saves time compared to the manual visual interpretation method, their detection performance depends on manually selected features or feature engineering. Furthermore, they exhibit unsatisfactory generalization performance, which further restricts their applications over wide areas \cite{stumpf2011object,prakash2020mapping}.
	
\begin{figure}[htbp]
		\setlength{\abovecaptionskip}{0.cm}
		\setlength{\belowcaptionskip}{-0.cm}	
		\centering
		\includegraphics[scale=0.3]{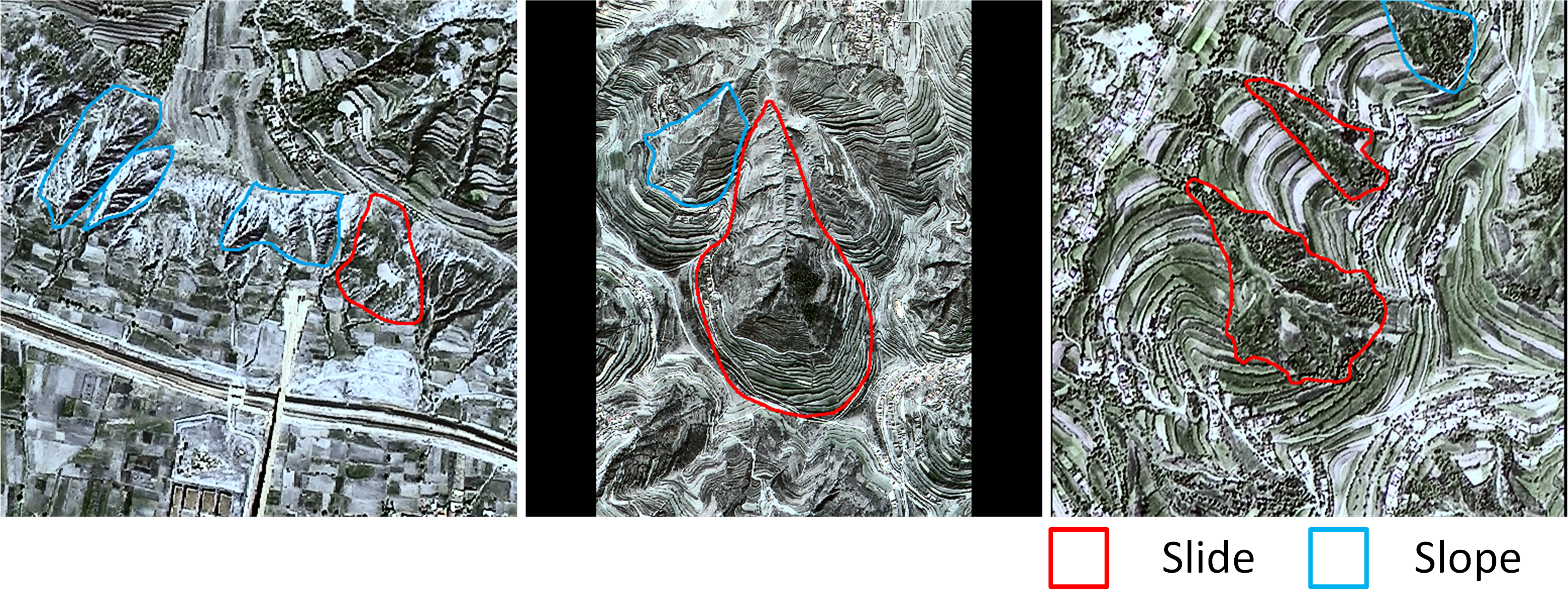}
		\DeclareGraphicsExtensions.
		\caption{Three old landslide samples in the Loess Plateau dataset to show the difficulty of recognizing old landslides from HRSI data.}
		\label{OldLandSlide}
	\end{figure}

	With the explosive growth of convolution neural network-based deep learning methods, semantic segmentation models gradually become mainstream in the field of landslide detection. From R-CNN \cite{he2017mask,girshick2015fast,ren2015faster} and FCN \cite{long2015fully}, through to the DeepLab series \cite{chen2018encoder}, semantic segmentation models have demonstrated excellent feature extraction and prediction capabilities in various fields. For automatic landslide detection, many advanced semantic segmentation networks have been proposed and validated to offer significant improvements in efficiency and  accuracy of large-scale detection \cite{yu2017landslide,fang2020gan,soares2020landslide,ji2020landslide,ju2020automatic,9428612,ghorbanzadeh2019optimizing}.
	However, there remain big challenges for reliable old landslide recognition using high-resolution remote sensing image (HRSI) data, due to the following challenges: 
	
	1) Visual blur problem of old landslides. As old landslides normally remain inactive for a long time, their morphology features have been partially or strongly transformed by natural or human activities\cite{landsildebook}. As a result, their color, tone, texture, and shadow features become gradually blurred, and their differences with surrounding slopes are severely reduced, which makes reliable detection so diffcult that even a geological expert may miss old landslides. We illustrate the visual blur between four landslides and their nearby slopes in Fig.~\ref{OldLandSlide}, where the red and blue curves  mark the landslide and slope, respectively; and 	
	
	2) Small-sized datasets. The reliable detection of visually blurred old landslides demands sufficient and well-labeled samples. However, the production of a high-quality landslide dataset from HRSI data is highly time- and labor-consuming, which requires significant professional expertise, leading to a scarcity of open large-scale datasets. Consequently, the size of available samples is far from enough to train high-complexity semantic segmentation models\cite{duin1995small}.
	
	In this paper, a novel iterative classification and semantic segmentation network (ICSSN) is proposed to detect visually blurred old landslides from a small-sized HRSI dataset. The proposed ICSSN utilizes a multiple-task learning-based iterative optimization method to enhance both object-level and pixel-level classification accuracy by iteratively training the feature extraction module shared by the object classification and semantic segmentation networks. Moreover, a sub-object-level contrastive learning (SOCL) paradigm is developed to extract salient features from the back and side walls of landslides, which contribute most to reliable landslide detection. The iterative optimization and SOCL can fully utilize small-sized datasets and distill semantic features more efficiently.  
	
	The main contributions of this paper are as follows:
	\begin{itemize}
		\item The ICSSN is proposed for visually blurred old landslide detection using HRSI data. The ICSSN greatly improves the reliability of both pixel-level and object-level classification by iteratively optimizing the weight-sharing feature extraction networks shared by the object classification and semantic segmentation tasks. The highly reliable object classification network can guide the feature extraction of a highly complex semantic segmentation network, while the semantic segmentation network can supplement the semantic features for object classification, thus improving both pixel-level and object-level classification performance;
		
		\item The SOCL method is developed in a semantic segmentation network for high-efficiency salient feature extraction. Motivated by the fact that side and back walls contribute most to old landslide detection, supervised contrastive learning is employed to extract salient features and the boundary information is utilized to construct sub-object-level positive samples, which is verified to be capable of distilling semantic features much more efficiently; and
		
		\item Extensive experiments are carried out to validate the efficacy of the proposed ICSSN. Our experimental results show that both the pixel-level and object-level classification performances are markedly improved, while the ablation study shows the iterative optimization strategy and the SOCL play important roles in improving the detection performance of our proposed ICSSN.
		
	\end{itemize}
	
	\begin{figure*}[hb]
		\centerline{\includegraphics[scale=0.35]{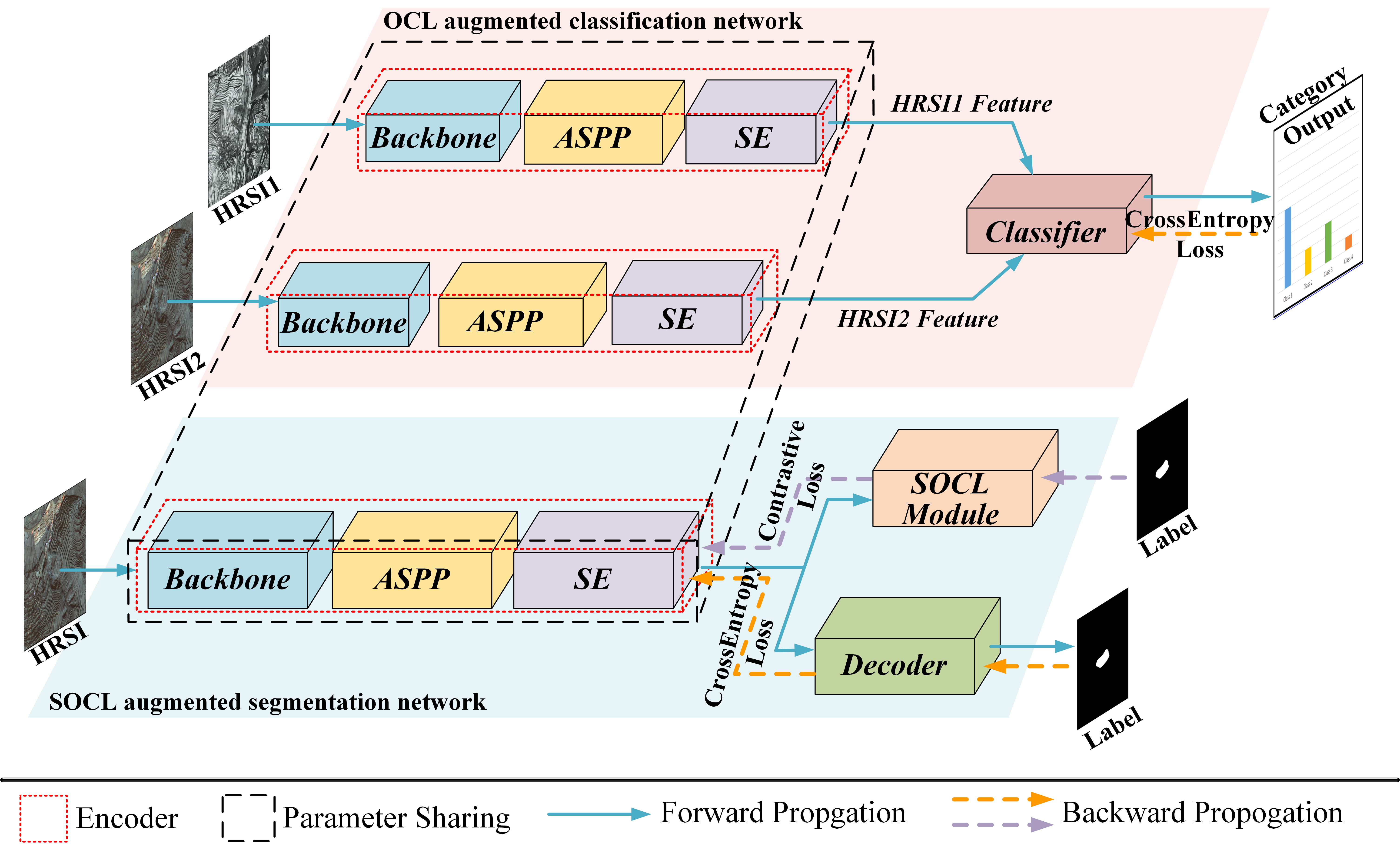}}
		\DeclareGraphicsExtensions.
		\caption{Architecture of ICSSN, including a classification network augmented by object-level contrastive learning and a semantic segmentation network augmented by sub-object-level contrastive learning.}
		\label{NetworkStructure}
	\end{figure*}

	\section{Related Work}
	The CNN-based deep learning methods can effectively extract semantic features, and thus have been widely applied to detecting landslides. 
	
	Semantic segmentation models have attracted the most attention in landslide detection from HRSI data. In \cite{fang2020gan}, a siamese convolutional neural network framework is proposed, where a generative adversarial network (GAN) is used to suppress the temporal difference between two HRSIs in different time phases, and a siamese network is employed for pixel-level change detection. Transformer network is introduced in\cite{tang2022automatic} to enlarge the  receptive field, while overlapping patch  introduced to ascertain the relationship between the associated patches, and then feature extraction efficiency is improved through the use of a simple MLP decoder. 
	
	Besides HRSI, other kinds of data, such as digital elevation model (DEM) and geological base data, can present useful information for landslide detection. Then, multi-source data fusion models have been investigated for landslide detection. In \cite{soares2020landslide}, the U-Net uses DEM data as an additional channel for HRSIs to provide terrain information such as the slope and aspect. The experimental results verify that introducing DEM data is conductive to enhancing the performance of new landslide detection. In \cite{ji2020landslide}, an attention mechanism is introduced into the CNN-based model, and DEM data is used as an additional channel to improve the detection accuracy of landslides.
	In \cite{huang2020deep}, geological base data has been utilized. By introducing an additional fully-connected autoencoder and sparse loss function, the nonlinear correlation problem of environmental factors is solved, and the feature expression ability of the model is improved, enabling unsupervised landslide recognition. 
	
	Although many deep learning-based works have been proposed, their research efforts focus mainly on the detection of visually distinct landslides whose optical features are significantly different from the background, with little attention given to visually blurred old landslides. In a recent study \cite{du2021landslide}, several mainstream semantic segmentation models are evaluated for old landslide detection and achieves the best F1 score of 0.417. In our previous work \cite{liu2023feature}, we design the FFS-Net model to extract semantic features from HRSI and DEM data, and then fuse the heterogeneous features in the semantic space to considerably enhance the landslide detection performance than the DeepLabv3+ scheme. DEM data can present more useful information for landslide detection, but also place a much higher requirement on the training dataset \cite{DSMFusion2019,MultiModal2019,Zeng2020}.
	
	This paper focuses on the challenging task of visually blurred old landslide detection using limited HRSI data. To address the problem caused by small-sized datasets, contrastive learning is exploited thanks to its effectiveness in feature extraction. Previous works have employed contrastive learning for both object classification and semantic segmentation. In \cite{wu2018unsupervised}, every instance and its data augmentation samples constitute the positive samples, while all the other samples constitute the negative ones. By comparing the similarities and differences of the contrastive samples, semantic features are extracted by a CNN and non-parameters softmax. A memory bank is introduced to save the features and the NCE loss function is employed to calculate loss and perform back propagation. Finally, a compact feature space and a good classification accuracy are obtained. The authors of  \cite{he2020momentum} construct a large and consistent dynamic dictionary with queueing and moving average encoders to facilitate unsupervised contrastive learning. The contrastive loss is further extended in \cite{khosla2020supervised} by designing a full-supervised contrastive loss so that it can effectively utilize label information. In \cite{wang2021exploring},  a pixel-wise contrastive algorithm is proposed for semantic segmentation in the fully supervised setting. Semantic segmentation is improved by the contrastive learning strategy, and the contrastive loss is computed for each pixel during the training process to assist in optimizing the feature space. 
	Building on these works, we improve the recognition reliability of old landslide from the following two aspects. 
	
	1) Inspiring by the expert knowledge from artificial visual interpretation approaches, the salient features for reliable old landslide detection mainly include:  1) the height drop characteristics between landslide and background, especially at the location of the back wall and the side wall; and 2) the tongue or horseshoe shape of the deposits~\cite{landsildebook}. In this paper, we designed a SOCL method to extract salient features from landslide side walls, back walls or deposits, which induces the semantic segmentation model to focus more on these parts that contribute the most to the recognition results and achieves more reliable semantic feature extraction; and
	
	2) Under the principle of model collaboration for multi-task learning\cite{zhang2018overview}, we designed an iterative optimizing method for feature extractor shared by object classification and semantic segmentation tasks. By sharing the feature extraction network, the global features of the landslide are extracted by the object classification network aided by OCL. Then the semantic segmentation network is instructed to extract locally salient features through SOCL, which in turn augments the object classification network with more comprehensive semantic features through iteration. Ultimately an overall improvement in target-level and pixel-level classification performance is achieved, which is verified by comparative and ablation experiments.

	\section{Proposed Method}
	
	We design a trainable network group and formulate a matched training algorithm to realize automatic old landslide detection. In what follows we first specify the overall architecture, details of each network of this network group, and finally the model training strategy.

	\subsection{Overall Structure}
	As shown in Fig.~\ref{NetworkStructure}, we partition the ICSSN into two separate branches, namely the upper branch for object classification and the lower branch for semantic segmentation. Both branches consist of an encoder and a classifier (decoder), which extract semantic features from RGB images and yeild object or pixel types, respectively. Through alternate training and iterative optimization, the upper branch guides the global feature extraction of the semantic segmentation network, while the lower branch extracts more abstract and comprehensive semantic features to improve the classification accuracy of the upper branch.
	
	\subsection{OCL Augmented Object Classification Network}
	
	We designed a classification network with the siamese structure to detect whether samples contain landslides by OCL.
	As shown in the top half of Fig.~\ref{NetworkStructure}, the classification network consists of a feature extraction module and a classifier module. 
	
	\subsubsection{Encoder}
	The feature extraction module adopts a weight-sharing siamese network architecture, which is composed of two encoders. Inspired by the classical DeepLabv3+ network structure, an SE (Squeeze-and-Excitation) attention module and an encoder-to-decoder skip connection structure are introduced at the encoder side. Each encoder consists of the backbone (ResNet101), ASPP (Atrous Spatial Pyramid Pooling), and SE modules for the extraction of high-dimensional features from the input RGB pictures. These modules are detailed as follows:

	\begin{itemize}
		\item The ResNet101 is employed as the backbone for feature extractionattributed to its proven effectiveness in feature extraction. It comprises four residual blocks, each of which has a different resolution of feature maps. After cross-validation, we found that concatenating the feature maps of layers 2 and 4 produces a final output that balances between abstract features and detailed information;	
		\item The ASPP module is inserted for multi-scale feature fusion. By using the dilation convolution with different receptive fields to sample the feature maps, the encoder can effectively take into account both small-scale detailed texture information and large-scale overall morphology information simultaneously.	
		\begin{figure}[h]
			\centering
			\includegraphics[scale=0.45]{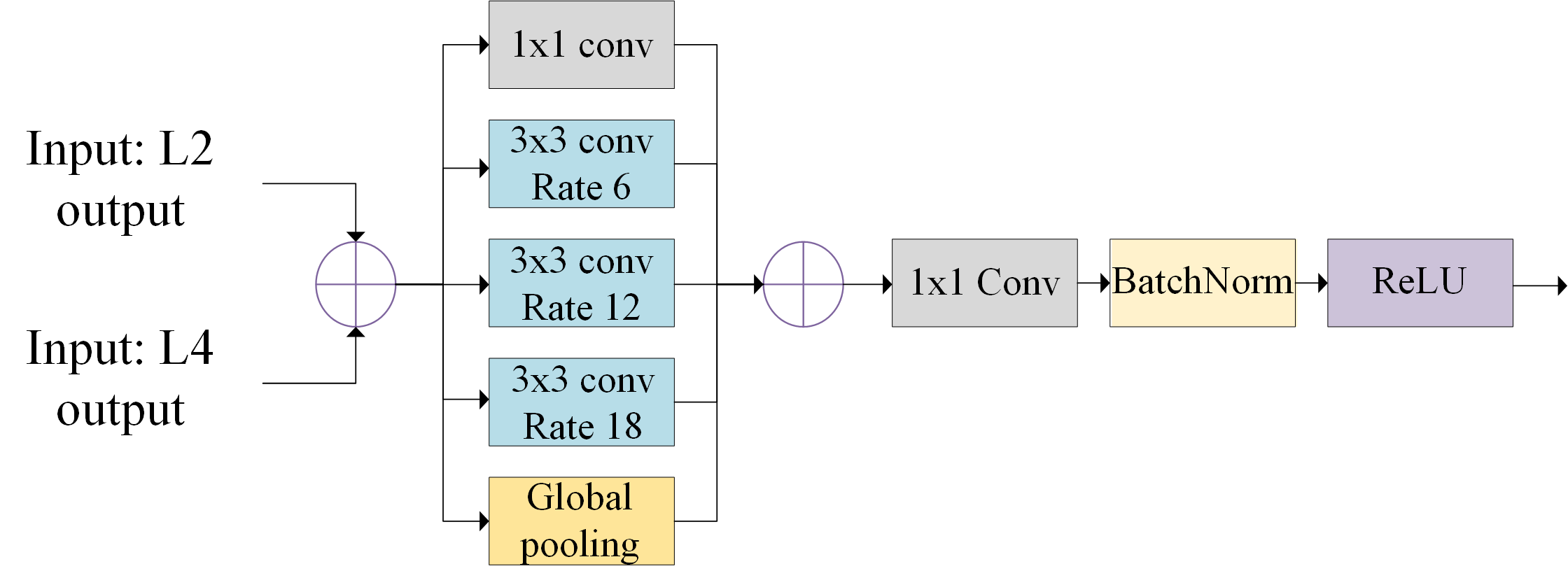}
			\DeclareGraphicsExtensions.
			\caption{ASPP structure.}
			\label{ASPP}
		\end{figure}
		\item SE module adaptively weights all channels in the feature map, allowing important channels to have greater weights and yielding a substantial enhancement in the capability to extract features.
		\begin{figure}[h]
			\centering
			\includegraphics[scale=0.4]{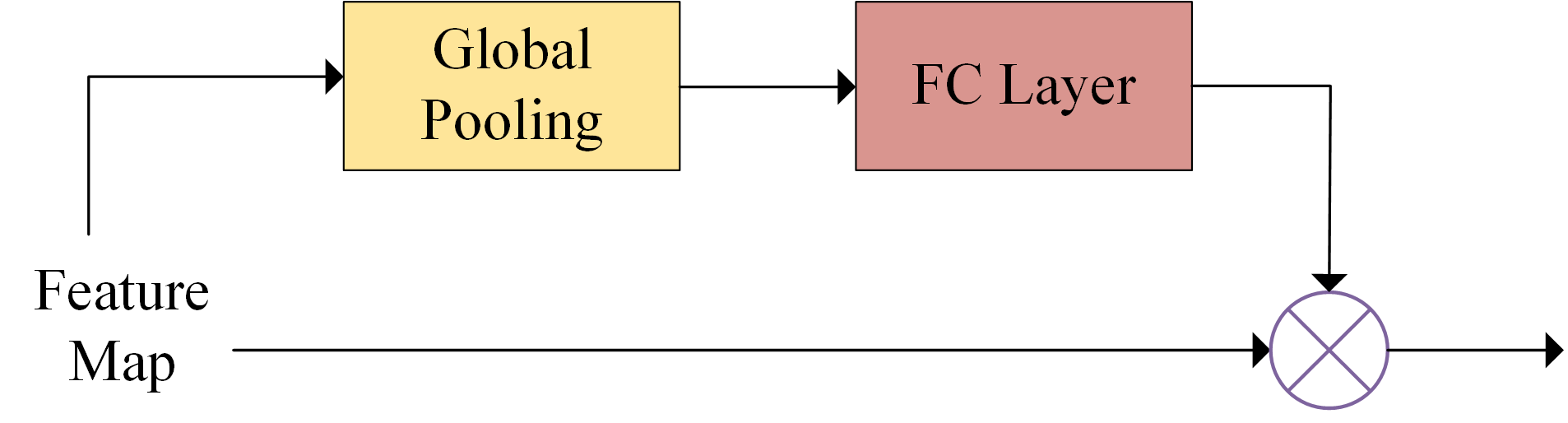}
			\DeclareGraphicsExtensions.
			\caption{SE structure.}
			\label{SE}
		\end{figure}
	\end{itemize}
	\subsubsection{Classifier}
	The classifier module consists of a global maximum pooling layer and two fully-connected layers, which maps and weights the extracted features to obtain the prediction probability for each category. The category with the highest prediction probability is yielded.
	
	Compared with classic classification networks, we employ a weight-sharing siamese network to improve the feature extraction capability and then concatenate the extracted global semantic features. This design offers the following two advantages. 
	\begin{figure}[h]
		\centering
		\includegraphics[scale=0.4]{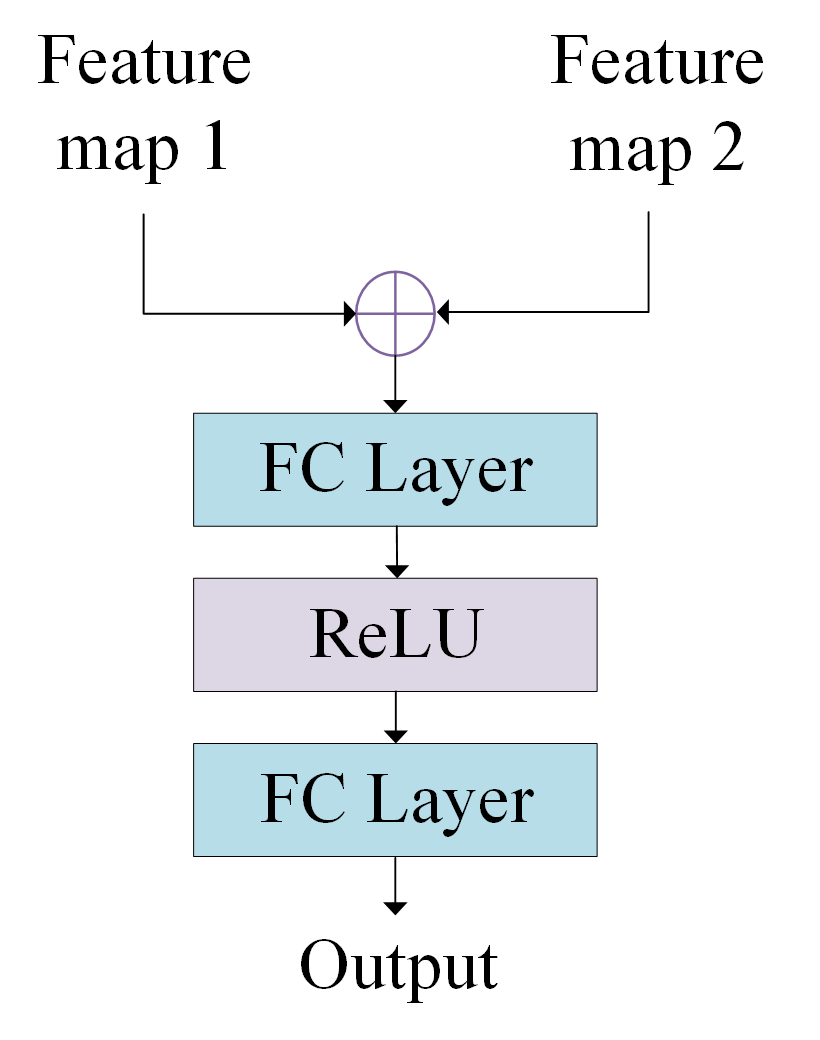}
		\DeclareGraphicsExtensions.
		\caption{Object-level contrastive-learning-based classification network.}
		\label{Classifier}
	\end{figure}
	Firstly, it allows the encoder to process two images simultaneously and concatenate the features, providing more features for the network to learn. Secondly, two input samples can form four distinct combinations for the classification prediction, i.e., landslide-landslide, landslide-slope, slope-landslide, and slope-slope, so that the network can learn the semantic features more efficiently by comparing the similarity and difference between the two samples. Specifically, when two samples are of the same category (positive-positive pair or negative-negative pair), the encoder learns the common features of this category. When the input samples fall into different categories (positive-negative pair or negative-positive pair), different features are learned for the corresponding categories.
	
	\subsection{SOCL Augmented Semantic Segmentation Network}
	
	As depicted in the bottom half of Fig.~\ref{NetworkStructure}, the semantic segmentation network is composed of three components, namely an encoder, an SOCL module, and a decoder. The SOCL module is a novel design that leverages \textit{a priori} expert knowledge and supervised contrastive learning to enhance the capability of salient semantic feature extraction. In addition, the decoder uses a learnable reverse convolution layer instead of the traditional linear interpolation layer to improve the pixel-level prediction accuracy.
	
	\subsubsection{Encoder}
	The encoder of the segmentation network has the same structure as the classification network, so as to facilitate alternate training;
	\subsubsection{Decoder}
	
\begin{figure}[h]
		\centering
		\includegraphics[scale=0.38]{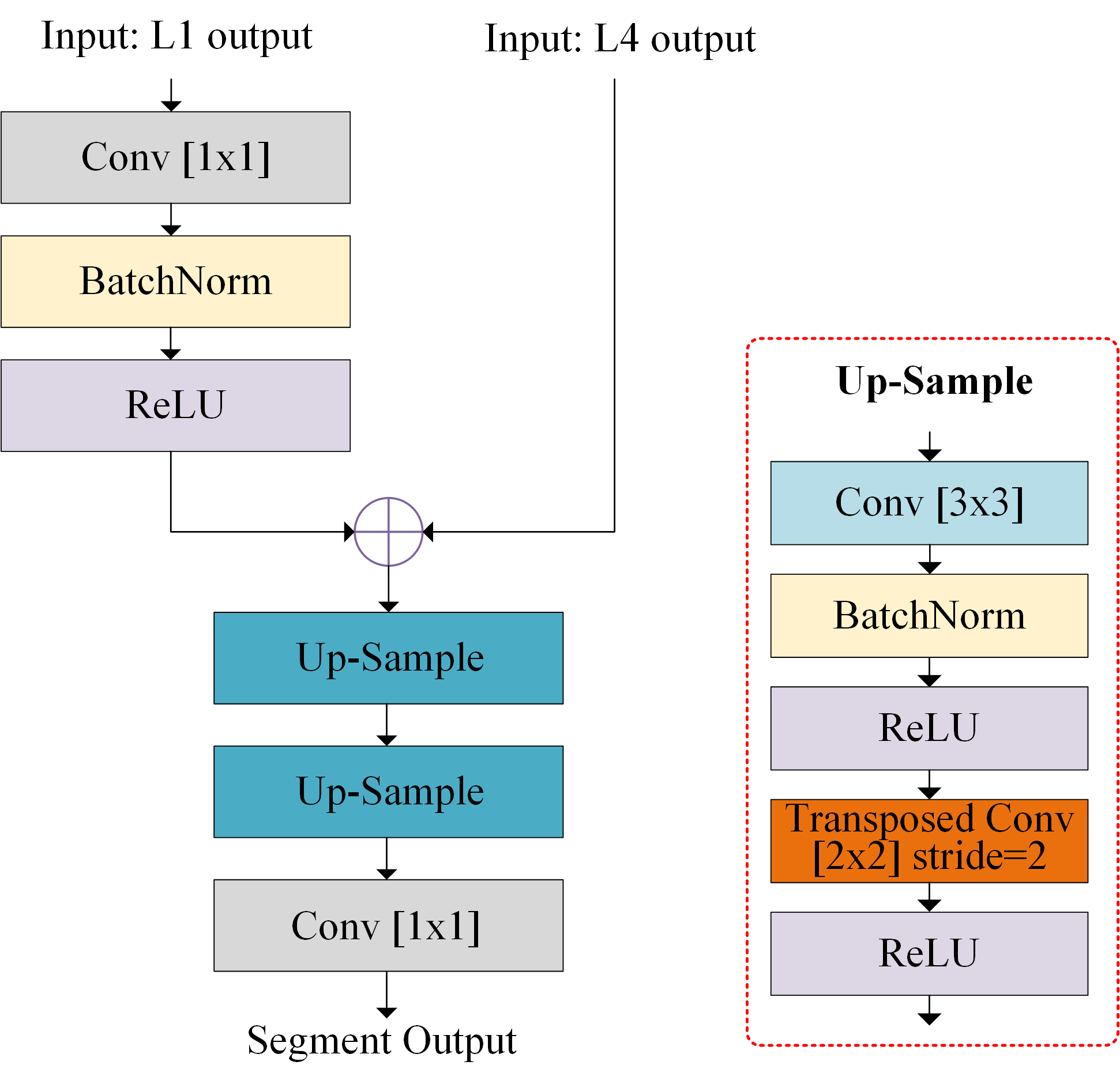}
		\DeclareGraphicsExtensions.
		\caption{Decoder architecture in the semantic segmentation network.}
		\label{Decoder}
	\end{figure}

	Based on the feature maps, the decoder classifies each pixel. We resort to our proposed decoder in \cite{liu2023feature}. As shown in Fig.~\ref{Decoder}, two transposed convolution layers are stacked to recover the resolution and avoid information redundancy to the greatest extent. In addition, the dropout layer is employed to avoid over-fitting. The batch normalization layer is used to restrict the data fluctuation range, while the ReLU activation function is employed to increase network sparsity and avoid gradient disappearance.
	
	\subsubsection{SOCL Module}
	
\begin{figure}[h]
		\centering
		\includegraphics[scale=0.38]{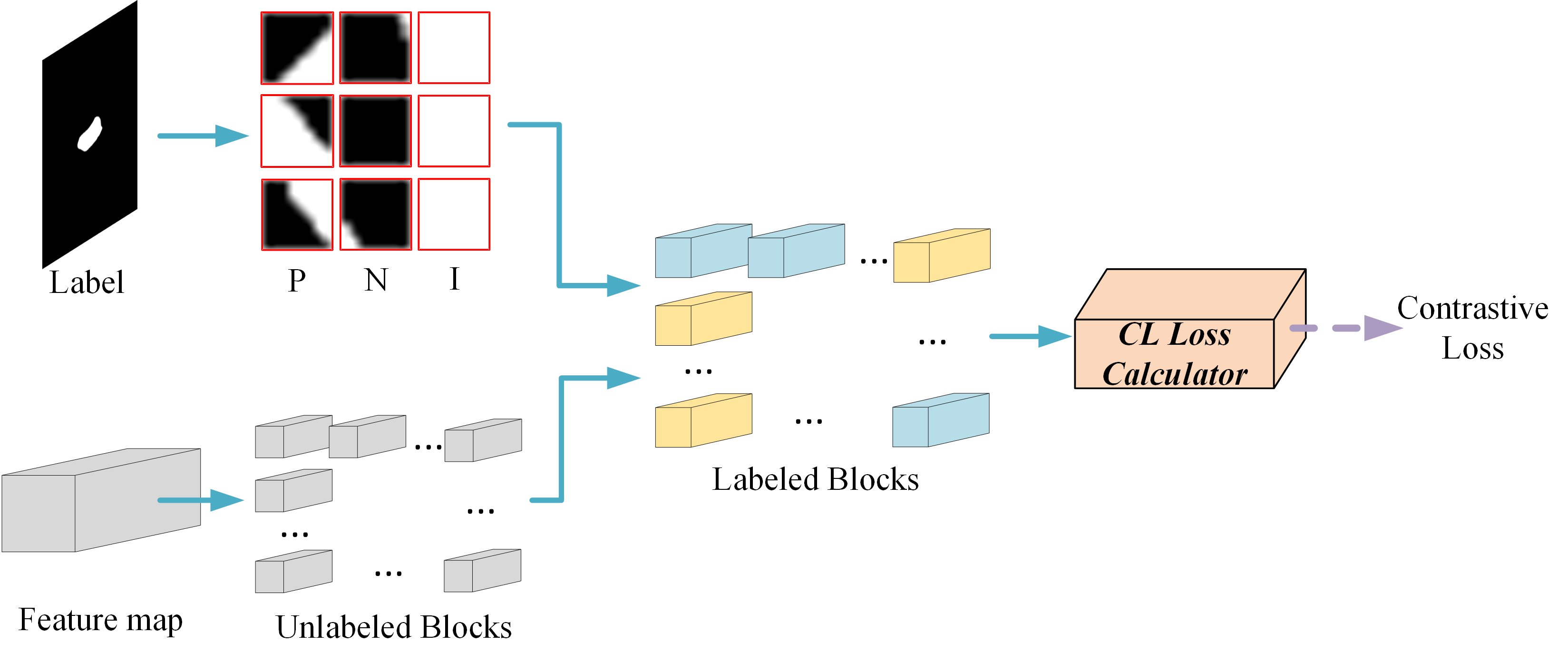}
		\DeclareGraphicsExtensions.
		\caption{SOCL module. In the SOCL labels, P represents the edge of the landslide, N represents the slope, and I represents the irrelevant class.}
		\label{SOCL}
	\end{figure}

	Fig.~\ref{SOCL} illustrates the SOCL module. The training of this module includes three steps, i.e., 1) using a frame of size $8 \times 8$ to filter the labels of the training images so as to obtain SOCL labels; 2) splitting the feature map in the spatial dimension, labeling it with SOCL labels and obtaining SOCL blocks; and 3) calculating the supervised contrastive loss by SOCL blocks.
	
	For the SOCL blocks of size $8\times8$, some are identified as the boundary of a landslide and then selected as the positive SOCL samples when they contain type-1 pixels (i.e., pixels located in a landslide) with a number lying between 7 to 57. When an SOCL block contain less than 7 type-1 pixels, it is regarded as background and treated as a negative SOCL sample. In particular, when an SOCL block contains more than 57 class-1 pixels, it is considered an internal part of a landslide and will be discarded due to its high similarity to slopes and thus a very limited contribution to the recognition of the landslide. This strategy is verified by the cross-validation experiment detailed in Section V. The selected positive, negative, and discarded SOCL samples are labeled as the 1, 0, and irrelevant classes, respectively. For each label $
	l \in \mathbb{R}^{H \times W \times 2}$, we obtain the corresponding label matrix $l^{'} \in \mathbb{R}^{\frac{H}{8} \times \frac{W}{8} \times 3}$.
	
	\begin{figure}[h]
		\centering
		\includegraphics[scale=0.55]{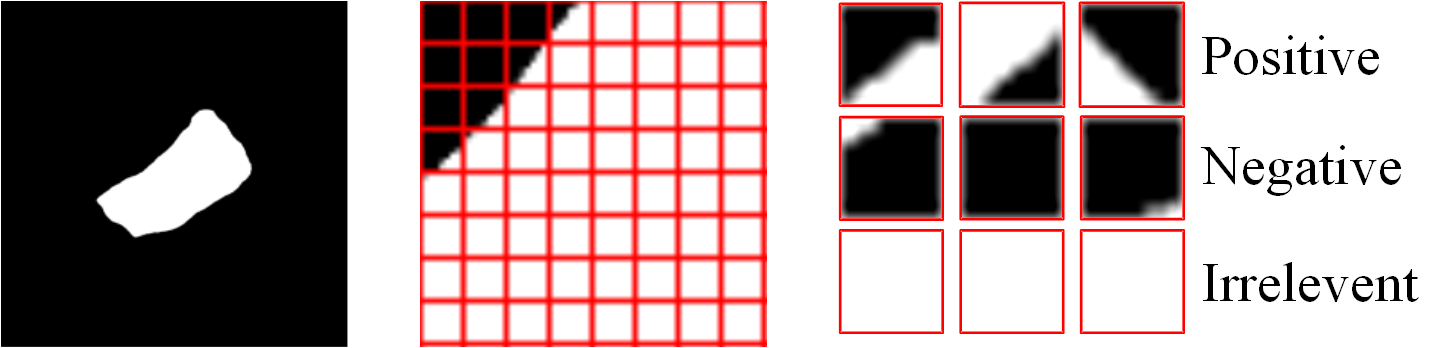}
		\DeclareGraphicsExtensions.
		\caption{The process of label filtering. Lables on the right are used in the SOCL module.}
		\label{Frame}
	\end{figure}
	
	In the training phase, we normalize the feature map yielded by the encoder $f \in \mathbb{R}^{\frac{H}{8} \times \frac{W}{8} \times C}$ in the spatial domain, and this step does not change the dimension of $f$. It should be noted that, $f$ and $l^{'}$ are of the same spatial resolution and can be aligned.
	
	Then we flatten the both $f$ and $l^{'}$ in the spatial domain and align features $f_{i}^{1 \times 1 \times C}\in f$ and labels $l_{i}^{' 1 \times 1 \times 3}\in l^{'}$ to form new sub-object-level feature-label pairs, each of which represents an $8\times8$ block in original samples. After that, we place all the SOCL pairs together to obtain $\frac{H}{8} \times \frac{W}{8} \times batch\_size$ pairs. Ignoring the pairs with labels of the irrelevant class, we randomly select a fixed number of pairs with classes 0 and 1 and then calculate their loss by the following supervised contrast learning loss function.
	
	\begin{footnotesize}
		\begin{equation}
			L_{i}^{C} = \frac{1}{\left| P_{i} \right|}{\sum\limits_{i^{+} \in P_{i}}{- {\rm log}\frac{{\rm exp}\left( i \cdot i^{+}/\tau \right)}{{\rm exp}\left( i \cdot i^{+}/\tau \right) + {\sum_{i^{-} \in N_{i}}{{\rm exp}\left( i \cdot i^{-}/\tau \right)}}}}},
		\end{equation}
	\end{footnotesize}where $P_{i}$ denotes the total number of positive SOCL samples, $i^{+}$ and $i^{-}$ represent the positive and negative SOCL samples, respectively. Meanwhile, the following cross-entropy loss function is employed to measure the difference between the predicted and real labels.
	
	\begin{equation}
		\left. L_{i}^{CE} = - y_{i}{\mathit{\log}(}\hat{y_{i}} \right) + \left( 1 - y_{i} \right)\left. {\mathit{\log}(}1 - \hat{y_{i}} \right),
	\end{equation}
	where $y_{i}$ indicates the label of pixel $i$ and, $\hat{y_{i}}$ indicates the logical output of this pixel. Therefore, the loss function of the semantic segmentation network is the combination of the two losses in (1) and (2).
	
	\begin{equation}
		L^{Seg} = {\sum_{i}} {\left(L_{i}^{CE} + \lambda L_{i}^{C}\right)},
	\end{equation}
	where $\lambda$ is the pre-set weight.
	
	\subsection{Training Strategy}
	
\begin{figure}[h]
		\centering
		\includegraphics[scale=0.7]{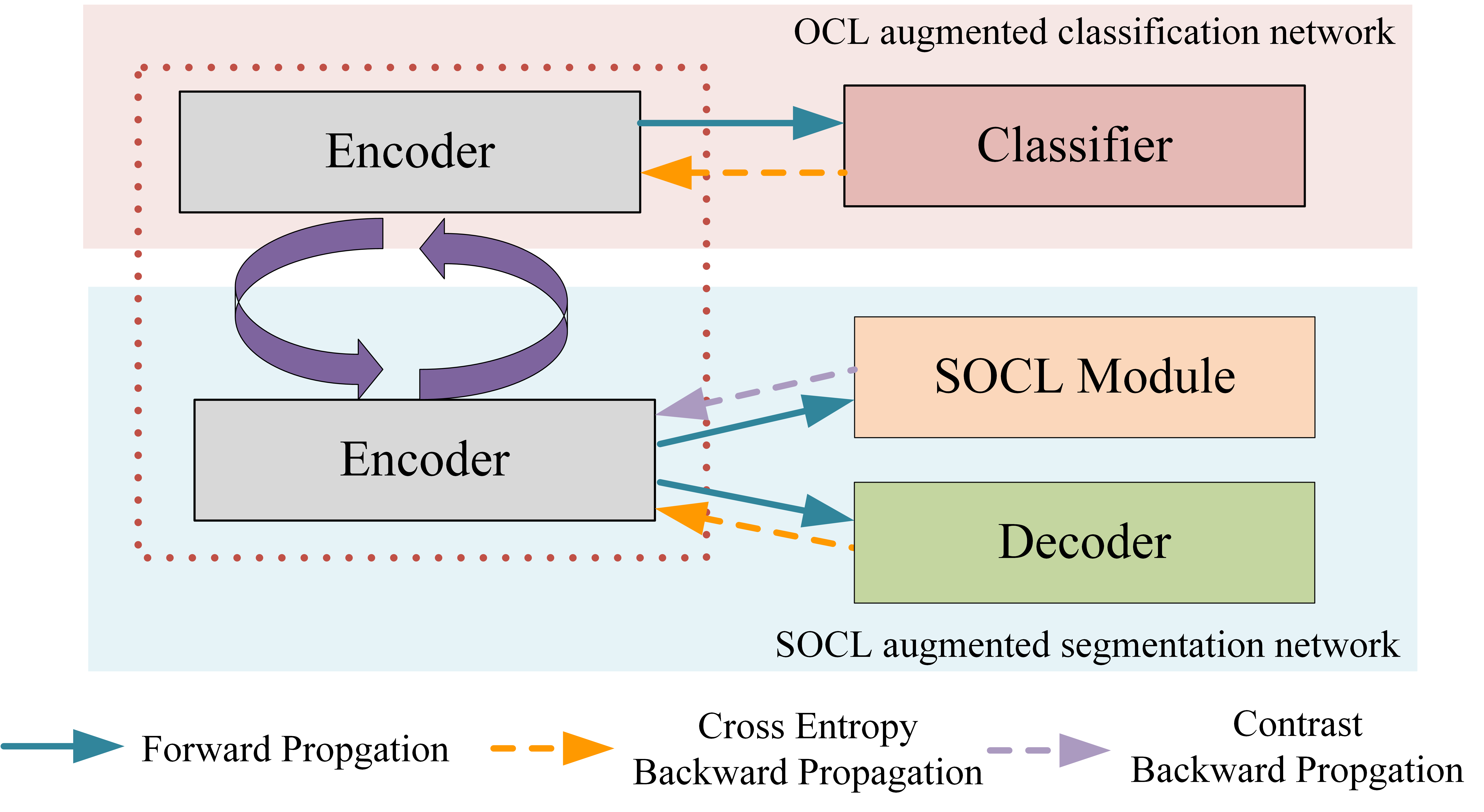}
		\DeclareGraphicsExtensions.
		\caption{Iterative training of the two networks, which share the same encoder.}
		\label{Iteration}
	\end{figure}

	A collaborative model training strategy is designed to iteratively train the two branches, and then continuously update the shared encoder. The iterative training process is illustrated in Fig.~\ref{Iteration}.
	
	The iterative training process alternates between the two networks. That is, the upper branch of the object classification network is trained until convergence, followed by the training of the lower branch of the semantic segmentation network until convergence. The whole training process is completed when both networks converge. During this alternation process, the encoder parameters in the previously converged network are used as the initial parameters for the latter network. The OCL augmented classification network is of high accuracy and can extract key features for object-level classification. Based on multi-task learning, it can effectively provide guidance and iterative calibration for the task of feature extraction of the SOCL augmented semantic segmentation network through the iterative mechanism. Generally, the object classification network extracts more abstract and global features with greater reliability due to its simpler task, while the semantic segmentation network extracts more local and detailed features with lower reliability due to the much harder task of pixel-level classification. Through iterative training, the encoder learns both local detailed features and global abstract features so that the two networks can realize the object-level and pixel-level classification, respectively. 
	
	Before each iteration, warm-up training is performed. The encoder parameters are frozen, and the decoder (classifier) is trained until convergence. After the warm-up training, the encoder parameters are unfrozen, and joint training is performed alongside the decoder (classifier). The warm-up training prevents feature mismatch between the optimized encoder of the previous network and the decoder (classifier) of the last iteration of the latter network and helps verify the sharing feature space. The training process is listed in Algorithm \ref{alg1}.

	\begin{algorithm}
		\renewcommand{\algorithmicrequire}{\textbf{Initialize:}}
		\renewcommand{\algorithmicensure}{\textbf{   }}
		\caption{Iterative Model Training}
		\label{alg1}
		\begin{algorithmic}[1]
			\REQUIRE 
			\ENSURE $N_{c}$: Classification network, containing $Encoder$ and $Classifier$
			\ENSURE $N_{s}$: Segmentation network, containing $Encoder$ and $Decoder$.
			\REPEAT
			\STATE Train $N_{c}$ until convergence.
			\STATE Load convergent $N_{c}$($Encoder$) parameters to $N_{s}$($Encoder$)
			\STATE Freeze $N_{s}$($Encoder$) and train $N_{s}$($Decoder$) until convergence.
			\STATE Unfreeze $N_{s}$($Encoder$) and train $N_{s}$ until convergence.
			\STATE Load convergent $N_{s}$($Encoder$) parameters to $N_{c}$($Encoder$)
			\STATE Freeze $N_{c}$($Encoder$) and train $N_{c}$($classifier$) until convergence.
			\STATE Unfreeze $N_{c}$($Encoder$) 
			\UNTIL 	Loss of $N_{c}$ and $N_{s}$ minimal.
		\end{algorithmic}  
	\end{algorithm}

	\section{Experiments}
	In this section, the proposed ICSSN is evaluated ion a real-world dataset. All experiments are conducted using PyTorch on two Nvidia 3090 GPUs, a 12th Gen Intel(R) Core(TM) i9-12900K, and 128 GiB of memory.
	
	\subsection{Experimental Dataset}
	
	The experiment zones are located in the northwest of China, which is in the transition zone between the western Qinling Mountains and the Longxi Loess Plateau. The soil parent material in this area is eluvial and accumulative, leading to soil with weak water permeability and erosion resistance. Sparse vegetation and heavy rainfall contribute to poor geological stability, which is conducive to the occurrence of landslides.
	
	The landslides in the study area occurred long time ago. The shape, color, and texture characteristics of landslides are of close resemblance to the surrounding environment. The landslide boundaries are fuzzy and the back walls of the landslides are visually blurred. To make matters worse, the main bodies of some landslides are transformed into farmland or residential areas, making landslide detection much more challenging. 
	
	\begin{figure}[h]
		\centering
		\includegraphics[scale=0.6]{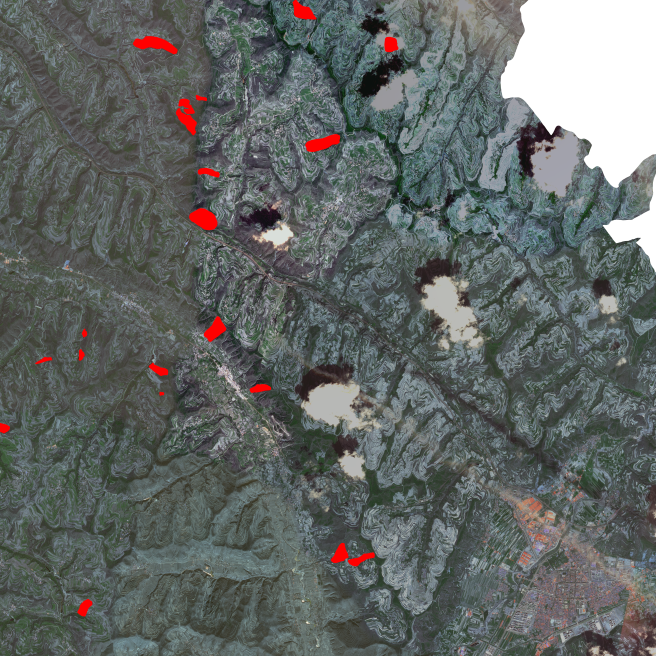}
		\DeclareGraphicsExtensions.
		\caption{HRSI landslide sample of one city. The red parts represent landslides.}
		\label{Oldlandslide}
	\end{figure}
	
	There are 304 landslide samples in this study area. The HRSI data have a resolution of 2 meters/pixel. Fig.~\ref{Oldlandslide} shows some landslide samples in one country.
	
	\subsection{Data Preprocessing}
	
	The original HRSIs are large in size. To facilitate model training, the images are resized into uniform samples of size $512 \times 512$. All the images containing landslides are selected to create positive samples after annotation. Images with no landslides are randomly collected to form negative samples.
	
	All the samples are divided into the training, validation, and test sets with a ratio of 6: 2: 2. Histogram equalization is applied to the entire dataset, and data augmentation techniques, including horizontal flipping, vertical flipping, and rotation, are used on the training and validation sets. The number of samples in each set is listed in Table~\ref{DatasetPartition}.
	
	\begin{table}[h]
		\centering
		\caption{ Datasets partition.}
		\label{DatasetPartition}
		\begin{tabular}{lccc}
			\hline
			& Training & Validation & Test \bigstrut\\
			\hline
			Slide & 1080  & 360   & 60 \bigstrut[t]\\
			
			Non-slide & 360   & 120   & 20 \bigstrut[b]\\
			\hline
		\end{tabular}%
	\end{table}
	
	\subsection{Performance Metrics}
	
	Both object-level classification metrics and pixel-level semantic segmentation metrics are used. The object-level classification metrics include pixel accuracy (PA), precision, recall, and F1-score, while the pixel-level classification metrics include pixel accuracy (PA), precision, recall, F1-score, and mIoU, which are commonly employed.
	
	\begin{small}
		\begin{equation}
			accuracy = \frac{TP + TN}{TP + FP + FN + FP} \,,
		\end{equation}
		\begin{equation}
			precision = \frac{TP}{TP + FP}\,,
		\end{equation}
		
		\begin{equation}
			recall = \frac{TP}{TP + FN}\,,
		\end{equation}
		
		\begin{equation}
			Landslide - IoU = \frac{TP}{TP + FP + FN}\,,
		\end{equation}
		
		\begin{equation}
			mIoU = \frac{1}{2} \times (class~1 - IoU + class~0 - IoU)\,,
		\end{equation}
		\begin{equation}
			F1 = 2 \times \frac{precision \times recall}{precision + recall}\,,
		\end{equation}
	\end{small}where\begin{small}
		$TP$, $TN$, $FP$ and $FN$
	\end{small}represent the true positive, true negative, false positive, and false negative at the pixel level, respectively.
	
	Additionally, an object-level classification metric can be obtained from the semantic segmentation metrics. Specifically, for a landslide sample, if more than 400 pixels within the landslide are accurately detected, it is considered  a correctly classified sample. For a slope sample, if less than 100 pixels are incorrectly classified as belonging to a landslide, it is considered a correctly classified slope sample. 
	
	\begin{equation}acc_+= \frac{1}{N_L}\sum_{i=1}^{N_L} I(C(\hat{P}_i \cap P_i)\geq 400) \,,
	\end{equation}
	
	\begin{equation}
		acc_- = \frac{1}{N_S}\sum_{j=1}^{N_S} I(C(\hat{F}_j \cup F_j)\leq 100)\,,
	\end{equation}
	where $N_L$ and $N_S$ are the numbers of landslide and slope samples in the test set, respectively, $\hat{P}_i$  and $\hat{F}_j$ are the predicted labels of the landslide and slope samples, respectively. $I(x)$ is the index function whose value is 1 when $x$ is true, or 0 when $x$ is false. $C(x)$ is a function to count the number of pixels with the same label in two comparison pictures.
	
	\subsection{Reference Models}
	Although several landslide detection models \cite{Ruijie2018} and heterogeneous information fusion models \cite{DSMFusion2019,MultiModal2019,Zeng2020} have been proposed, unfortunately, none of them has yet published the source code of their models. As a result, we cannot reproduce their results for comparison purposes. Furthermore, we cannot directly compare our model with the heterogeneous information fusion models \cite{DSMFusion2019,MultiModal2019,Zeng2020}  for three additional reasons: 1) the target of these models are not landslides; and 2) aerial images, DSM or SAR are used by these models, which are very different from HRSI data. Consequently, we have to evaluate the proposed model through comparative experiments with the baseline model and ablation experiments.
	
	\begin{figure}[h]
		\centering
		\includegraphics[scale=0.3]{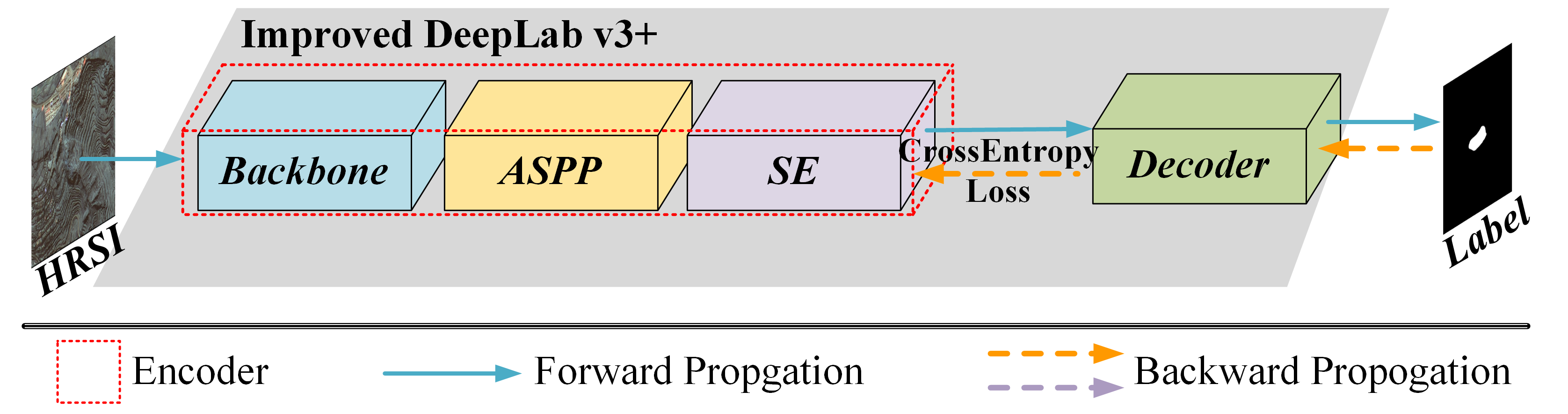}
		\DeclareGraphicsExtensions.
		\caption{Architecture of DeepLabv3+}
		\label{deeplab}
	\end{figure}
	
	Four comparative experiments are designed: 1) Improved Deeplabv3+\cite{liu2023feature} as the baseline model, whose architecture is shown in Fig.~\ref{deeplab};
	2) segmentation network augmented by SOCL; 3) object classification network augmented by OCL; and 4) the proposed ICSSN. 
	The hyper-parameters of model training are the same for all the four comparison models, which are listed in Table~\ref{Hyperparameters}.
	\begin{figure*}[htbp]
		\centerline{\includegraphics[scale=0.7]{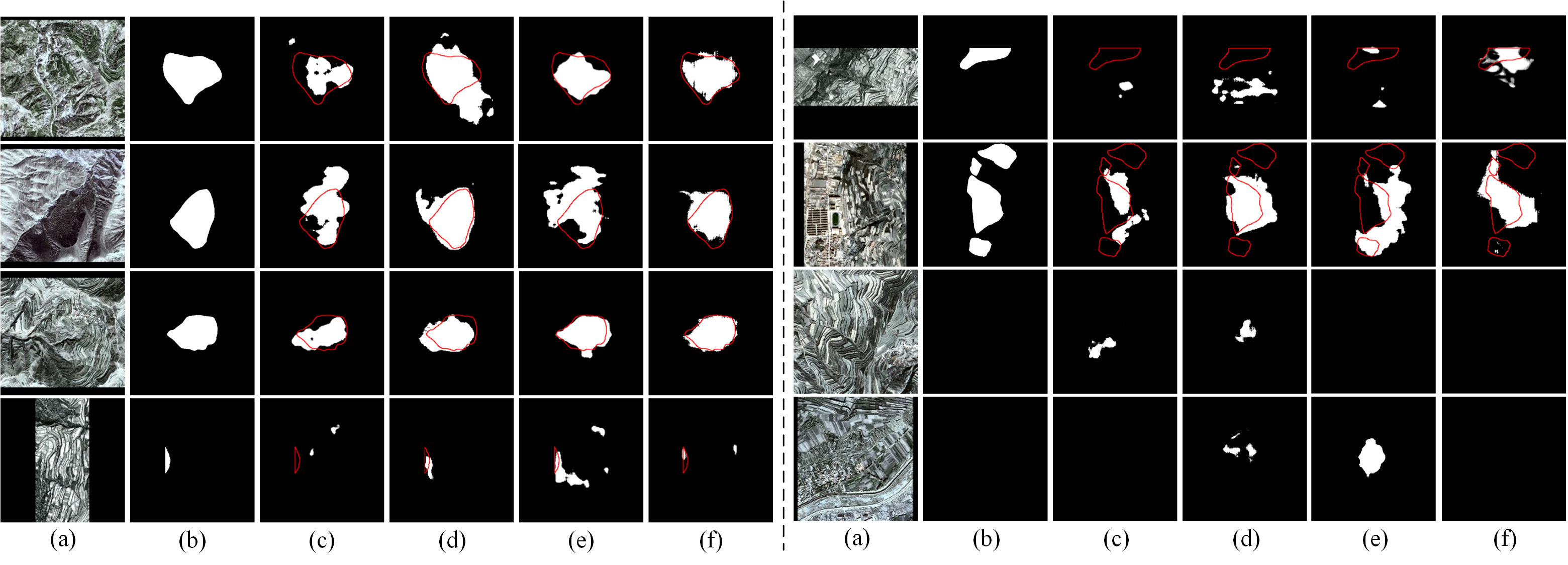}}
		\DeclareGraphicsExtensions.
		\caption{Visual results of segmentation on old landslides. (a): RGB image; (b): label; (c)-(f): Baseline; SOCL; OCL; ICSSN.}
		\label{ComResult}
	\end{figure*}
	\begin{table*}[h]
		\centering
		\caption{Pixel-level quantitative experimental results of segmentation networks.}
		\begin{tabular}{lcccccc}
			\hline
			\multicolumn{1}{c}{Method} & PA    & Precision & Recall &  Landslide-IoU & mIoU  & F1 \bigstrut\\
			\hline
			Baseline & 0.9445 & 0.4226 & 0.6284 &  0.3381 & 0.6405 & 0.5054 \bigstrut[t]\\
			SOCL & 0.9406 & 0.5176 & 0.5621 &  0.3688 & 0.6536 & 0.5387 \\
			OCL & 0.9431 & 0.4896 & 0.5911 &  0.3658 & 0.6535 & 0.5356 \\
			ICSSN & \textbf{0.9493} & 0.4531 & 0.6832 &  \textbf{0.3743} & \textbf{0.6610} & \textbf{0.5448} \bigstrut[b]\\
			\hline
		\end{tabular}%
		\label{PixelResultOfSeg}%
	\end{table*}
	\begin{table}[h]
		\centering
		\caption{ Model hyperparameters.}
		\label{Hyperparameters}
		\begin{tabular}{lcc}
			\hline
			& Classification & Segmentation \bigstrut\\
			\hline
			Number of workers & 4 & 4 \bigstrut[t]\\
			
			Batchsize & 4 & 4 \\
			
			Optimizer  & SGD & SGD \\
			
			Momentum & 0.9 & 0.9 \\
			
			Initial learning rate & 0.001 & 0.007 \\
			
			Weight decay & 0.0005 &0.0005 \\
			
			Learning rate decay strategy & cosine annealing & cosine annealing\\
			
			Epoch & 50    & 100 \bigstrut[b]\\
			\hline
		\end{tabular}%
	\end{table}
	
	The SOCL model employs the cross entropy as the segmentation loss function, which is defined as follows.
	\begin{equation}
		\left. Loss = - y_{i}{\mathit{\log}(}\hat{y_{i}} \right) + \left( 1 - y_{i} \right)\left. {\mathit{\log}(}1 - \hat{y_{i}} \right),
	\end{equation}
	where $y_{i}$ indicates the ground truth and $\hat{y_{i}}$ denotes the predicted mask.
	
	\section{Results and Discussions}
	In this section, we present experimental results with visualization as well as some detailed discussions on the obtained results. 
	
	\subsection{Comparative Experiments}
	In this subsection, we will show and discuss the experimental results achieved by the four comparison models on the same old landslide dataset.
	
	\begin{table}[h]
		\centering
		\caption{Object-level classification experimental results of the segmentation networks.}
		\begin{tabular}{lccc}
			\hline
			\multicolumn{1}{c}{Method} & Slope Acc & Landslide Acc & Avg. Acc \bigstrut\\
			\hline
			
			Baseline & 0.78  & 0.55  & 0.67 \bigstrut[t]\\
			SOCL & 0.78  & 0.55  & 0.67 \\
			OCL & 0.78  & 0.65  & 0.72 \\
			ICSSN & 0.78  & \textbf{0.90} & \textbf{0.84} \bigstrut[b]\\
			
			\hline
		\end{tabular}%
		\label{ObjectResultOfSeg}%
	\end{table}
	
	\subsubsection{Semantic Segmentation Results}
	The pixel-level classification experimental results are listed in Table~\ref{PixelResultOfSeg}, while the object-level classification results, which are derived from the segmentation network, are listed in Table~\ref{ObjectResultOfSeg}. The following observation can be made from the results in these two tables:
	
	\begin{itemize}
		\item The proposed ISCNN model achieves the best performance among all the comparison models. Specifically, compared to the baseline model, the F1 score increases from 0.5054 to 0.5448, the mIoU improves from 0.6405 to 0.6610, and especially the landslide-IoU is up from 0.3381 to 0.3743. Furthermore, the object-level accuracy of landslide classification is significantly enhanced from 0.55 to 0.9. These improvements clearly demonstrate the advantage of the proposed model in both object-level and pixel-level classification of old landslides;
		
		\item The effectiveness of the proposed SOCL module in improving boundary detection accuracy is demonstrated by the significant improvements in both pixel-level and object-level metrics when compared to the baseline model. Although the object-level accuracy of both classes remains the same as the baseline model, the SOCL network still achieves notable pixel-level improvements, with the F1 score increased from 0.5054 to 0.5387, the mIoU improved from 0.6405 to 0.6536, and the IoU of class 1 up from 0.3381 to 0.3688; and
		
		\item Regarding the object-level results of the segmentation network, the ISCNN model is superior, which increases from 0.55 to 0.65. As a result, the pixel-level performance is improved correspondingly. The F1 score increases from 0.5054 to 0.5356, the mIoU improves from 0.6405 to 0.6535, and the IoU of class 1 is up from 0.3381 to 0.3658.	
		
	\end{itemize}
	
	Some experimental results are also visualized in Fig.~\ref{ComResult}. As can be seen from the figure, the SOCL scheme predicts more accurate shapes, especially the boundary parts, when compared with the baseline scheme. The OCL scheme has fewer misjudgments and omissions, while the ISCNN model, which fuses both the SOCL and OCL schemes, is most accurate in both object-level classification and pixel-level classification. 
	
	\begin{figure}[h]
		\centering
		\includegraphics[scale=0.55]{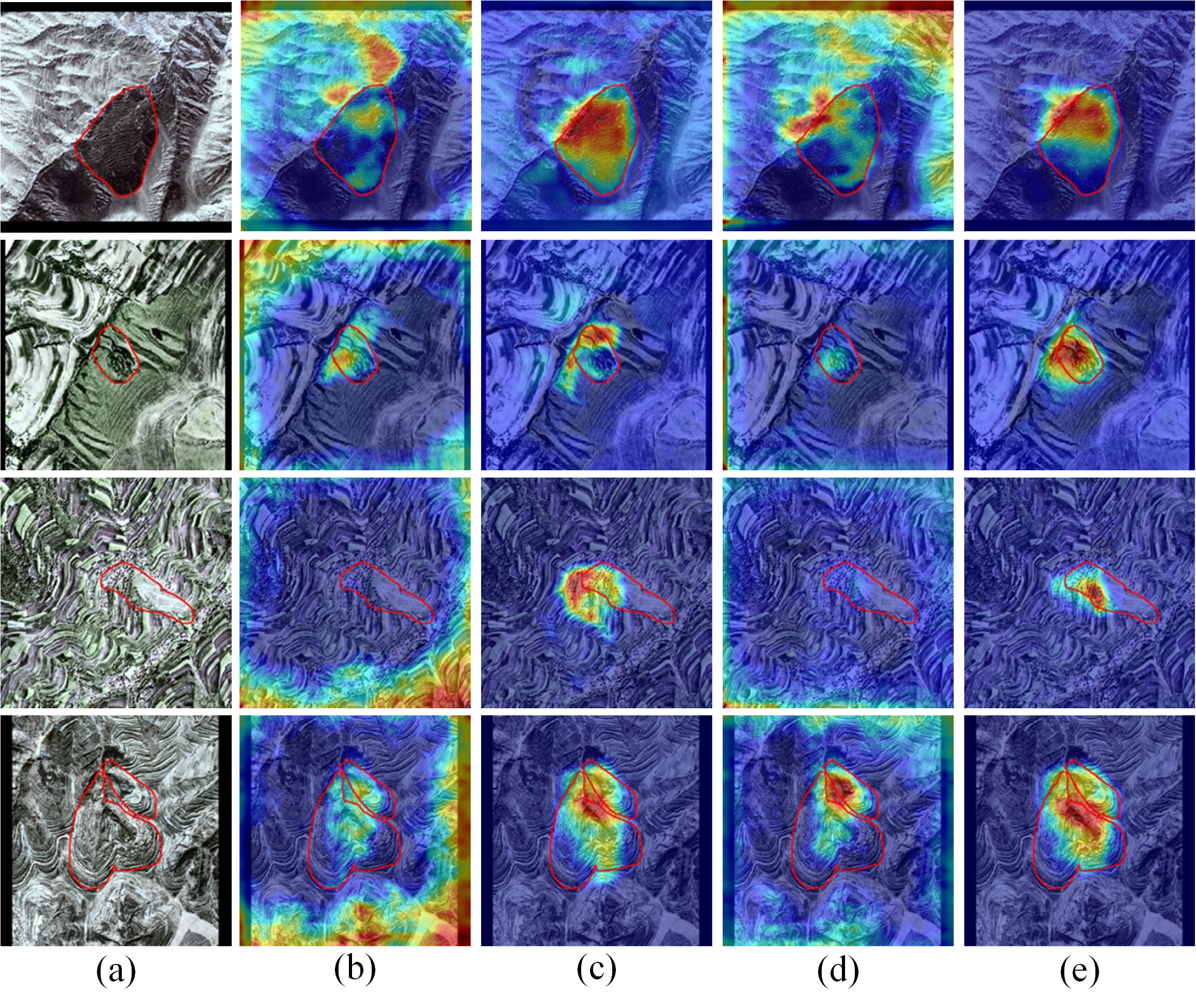}
		\DeclareGraphicsExtensions.
		\caption{Visual results by Grad-cam of the segmentation networks. (a): img; (b)-(e): Baseline; SOCL; OCL; ICSSN.}
		\label{SegHeatmap}
	\end{figure}
	
	The heat map analysis can effectively validate the design of a model, and we have employed the popular Gradient-weighted Class Activation Mapping (Grad-CAM) algorithm \cite{selvaraju2017grad} as our visualization method. The Grad-CAM method visualizes the contribution of each region of the input image to the predicted result by weighting the feature map and overlaying it with the original image. Fig.~\ref{SegHeatmap} visualizes the final layer of the encoder of our model. The heat maps demonstrate that the crucial areas in the heat map of the two schemes with the SOCL strategy are closer to the back and side walls, which verifies the effectiveness of the SOCL sample selection strategy and the SOCL module for detecting old landslides.

	\subsubsection{Object Classification Results}
	In both the OCL and ICSSN schemes, we introduced the classification networks. Table~\ref{classResult} lists the numerical results of these three classification schemes, i.e., no iteration, OCL object classification network, and ICSSN object classification network. Note that Tables \ref{classResult} and \ref{ObjectResultOfSeg} both present object-level performances, but they are derived from different networks. Table~\ref{classResult} is derived from the classification network, while Table~\ref{ObjectResultOfSeg} is derived from the segmentation network.
	\begin{table}[h]
		\centering
		\caption{Quantitative Classification Experimental results of the classification networks.}
		\begin{tabular}{lcccc}
			\hline
			\multicolumn{1}{c}{Method} & Accuracy & Precision & Recall & F1 \bigstrut\\
			\hline
			No iteration & 0.8375 & 0.9607 & 0.8167 & 0.8846 \bigstrut[t]\\
			OCL & 0.8625 & 0.9455 & 0.8667 & 0.9044 \\
			ICSSN  & \textbf{0.8875} & 0.9473 & 0.9003 & \textbf{0.9230} \bigstrut[b]\\
			\hline
		\end{tabular}%
		\label{classResult}%
	\end{table}
	
	As can be seen from the result that: 
	\begin{itemize}
		\item The ICSSN achieves the best F1 score followed by the OCL, and both are better than the no iteration scheme. This shows that the segmentation network can provide effective guidance to the classification network and strength the feature extraction ability of the encoder by using the collaborative model training strategy. This guidance is still effective when the segmentation network has better abilities;
		\begin{figure}[h]
			\centering
			\includegraphics[scale=0.55]{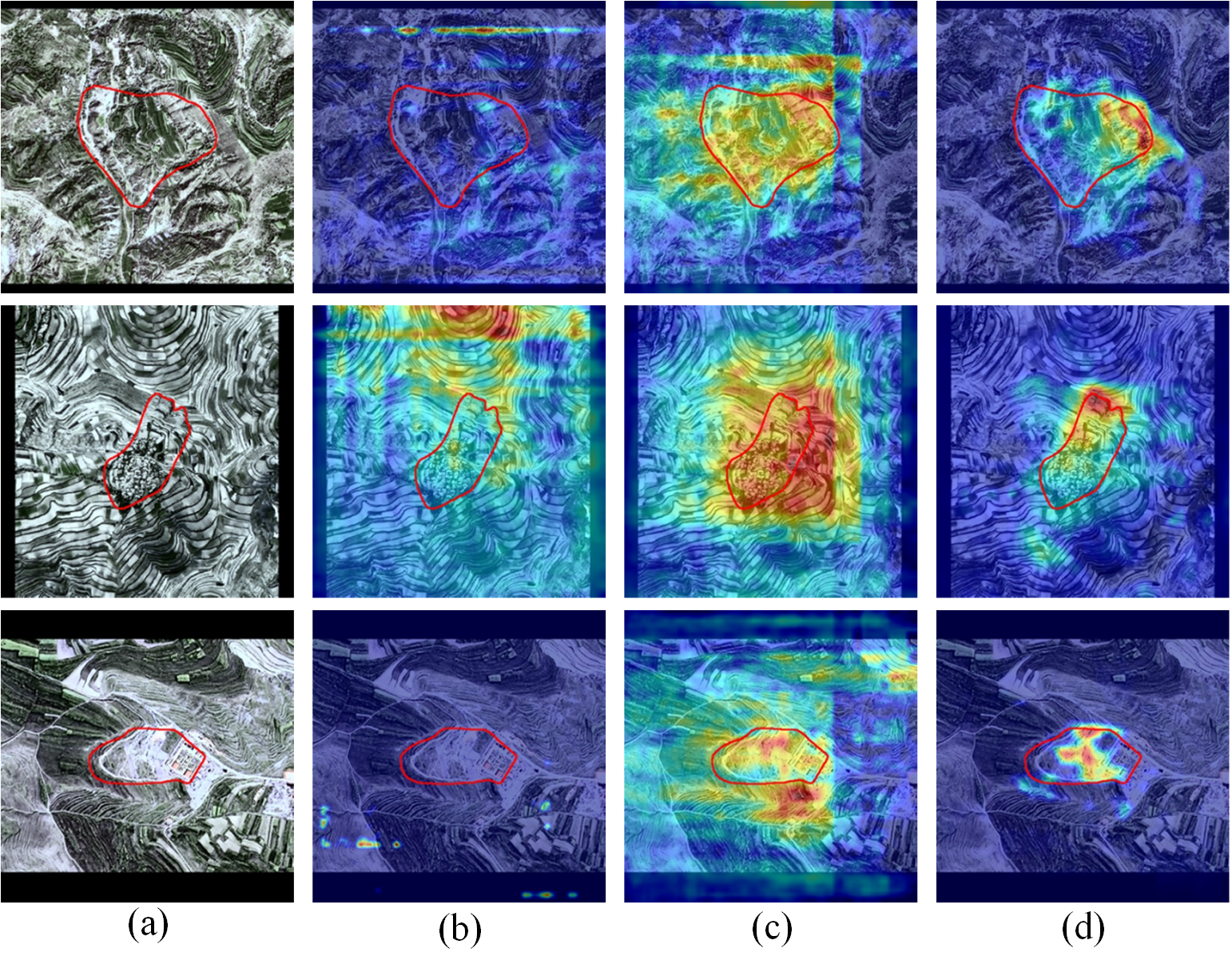}
			\DeclareGraphicsExtensions.
			\caption{Visual results by Grad-cam of the classification networks. (a): img; (b)-(d): no iteration; OCL; ICSSN.}
			\label{ClassHeatmap}
		\end{figure}
		\item Fig.~\ref{ClassHeatmap} visualizes the heat map of the classification network. It can be observed that the features extracted from the initial classification are not intuitive, and the prediction results are based on the entire image. Through iterative training with the segmentation network, the important area gradually shifts to the surroundings of the landslide, thereby improving the detection accuracy. Based on the above analysis, it is verified that the iterative training strategy is conductive to fusing the feature spaces of the classification network and the segmentation network, thus improving the performance of both branches.

	\end{itemize}
	
	\subsection{Cross-validation}
	
	In order to obtain the best performance, we optimize the design of the proposed model via cross-validation experiments.
	\begin{table}[h]
		\centering
		\caption{Cross-validation of the SOCL blocks selection strategy.}
		\begin{tabular}{lccc}
			\hline
			Strategy& Landslide-IoU & mIoU & F1 \bigstrut\\
			\hline
			Center  & 0.3427 & 0.6399 & 0.5105 \bigstrut[t]\\
			\textbf{Edge}  & \textbf{0.3688} & \textbf{0.6536} & \textbf{0.5387} \\
			Hybrid  & 0.3606 & 0.6482 & 0.5300 \bigstrut[b]\\
			\hline
		\end{tabular}%
		\label{PointPlan}%
	\end{table}
	\subsubsection{SOCL blocks selection strategy}
	We devise three strategies for selecting the SOCL blocks: 1) central strategy - the SOCL blocks are selected from the  the central parts of the landslides and they contain more than 57 class-1 pixels; 2) edge strategy - the SOCL blocks are chosen exclusively from the boundaries of the landslides and they contain class-1 pixels between 7 and 57; and 3) hybrid strategy - the SOCL blocks are randomly sampled from both the central and boundary parts of the landslides with equal probability.
	
	The experimental results presented in Table~\ref{PointPlan} indicate that the edge strategy performs the best, while the center strategy performs the worst. This observation confirms that the boundaries of a landslide contain more information relevant to landslide detection.
	\begin{table}[h]
		\centering
		\caption{Cross-validation of classifier.}
		\begin{tabular}{lcccc}
			\hline
			Method & Accuracy & Precision & Recall & F1 \bigstrut\\
			\hline
			Single-branch & 0.8250 & 0.9432 & 0.8167 & 0.8750 \bigstrut[t]\\
			Max Pool \& $1\times 1$ conv & 0.7375 & 0.9149 & 0.7167 & 0.8037 \\
			Max Pool \& $1*$FC & 0.8000   & 0.9400  & 0.7833 & 0.8545 \\
			Avg Pool \& $2*$FC & 0.7125 & 0.8776 & 0.7167 & 0.7889 \\
			\textbf{Max Pool \& $2*$FC} & \textbf{0.8375} & 0.9607 & 0.8167 & \textbf{0.8846} \\
			Max Pool \& $3*$FC & 0.7500  & 0.9000   & 0.7500  & 0.8182 \bigstrut[b]\\
			\hline
		\end{tabular}%
		\label{classifierVal}%
	\end{table}
	\subsubsection{Classifier}
\begin{figure}[h]
		\centering
		\includegraphics[scale=0.55]{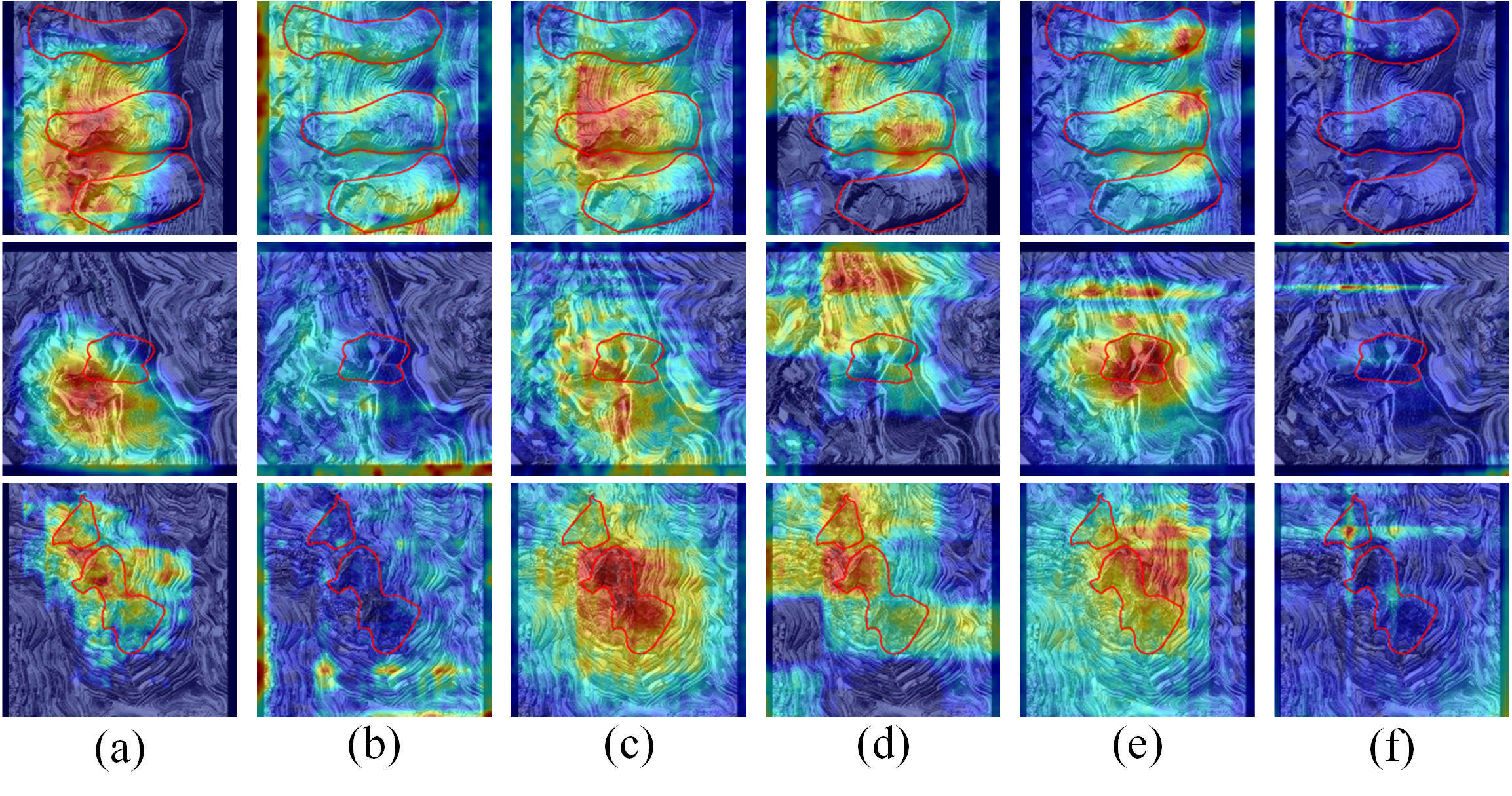}
		\DeclareGraphicsExtensions.
		\caption{Visual results by classifier cross-validation. (a)-(f) corresponds to six schemes respectively.}
		\label{ClassifierHeatmap}
	\end{figure}

	We design several classification networks, including the single-branch and two-branch classification networks with five types of classifiers. The numerical and visual experimental results are listed in Table~\ref{classifierVal} and Fig.~\ref{ClassifierHeatmap}, respectively, and both results show that the Max pool and $2*$FC schemes are preferred.
	
	\begin{table}[h]
		\centering
		\caption{Quantitative experimental results of the classification network during iteration.}
		\begin{tabular}{lcccc}
			\hline
			\multicolumn{1}{c}{Method} & Accuracy & Precision & Recall & F1 \bigstrut\\
			\hline
			First round & 0.8375 & 0.9607 & 0.8167 & 0.8846 \bigstrut[t]\\
			Second warm-up & 0.8250 & 0.9423 & 0.8167   & 0.8749 \\
			\textbf{Second round} & \textbf{0.8875} & \textbf{0.9473} & \textbf{0.9000}   & \textbf{0.9230} \\
			Third warm-up & 0.8625 & 0.9623 & 0.8500   & 0.9027 \\
			Third round & 0.8750 & 0.9629 & 0.8667   & 0.9122 \bigstrut[b]\\
			\hline
		\end{tabular}%
		\label{ClassificationIteration}%
	\end{table}
	
	\begin{table}[h]
		\centering
		\caption{Quantitative experimental results of the segmentation network during iteration.}
		\begin{tabular}{lccc}
			\hline
			& Landslide-IoU & mIoU & F1 \bigstrut\\
			\hline
			First warm-up  & 0.2403 & 0.5880 & 0.3875 \bigstrut[t]\\
			First round  & 0.3724 & 0.6567 & 0.5419 \\
			Second warm-up  & 0.3690 & 0.6551 & 0.5391 \\
			\textbf{Second round}  & \textbf{0.3743} & \textbf{0.6610} & \textbf{0.5448} \\
			Third warm-up  & 0.3652 & 0.6516 & 0.5350 \\
			Third round & 0.3715 & 0.6566 & 0.5418 \bigstrut[b]\\
			\hline
		\end{tabular}%
		\label{SegmentationIterationResult}%
	\end{table}
	\subsubsection{Iteration process}
\begin{figure}[h]
		\centering
		\includegraphics[scale=0.55]{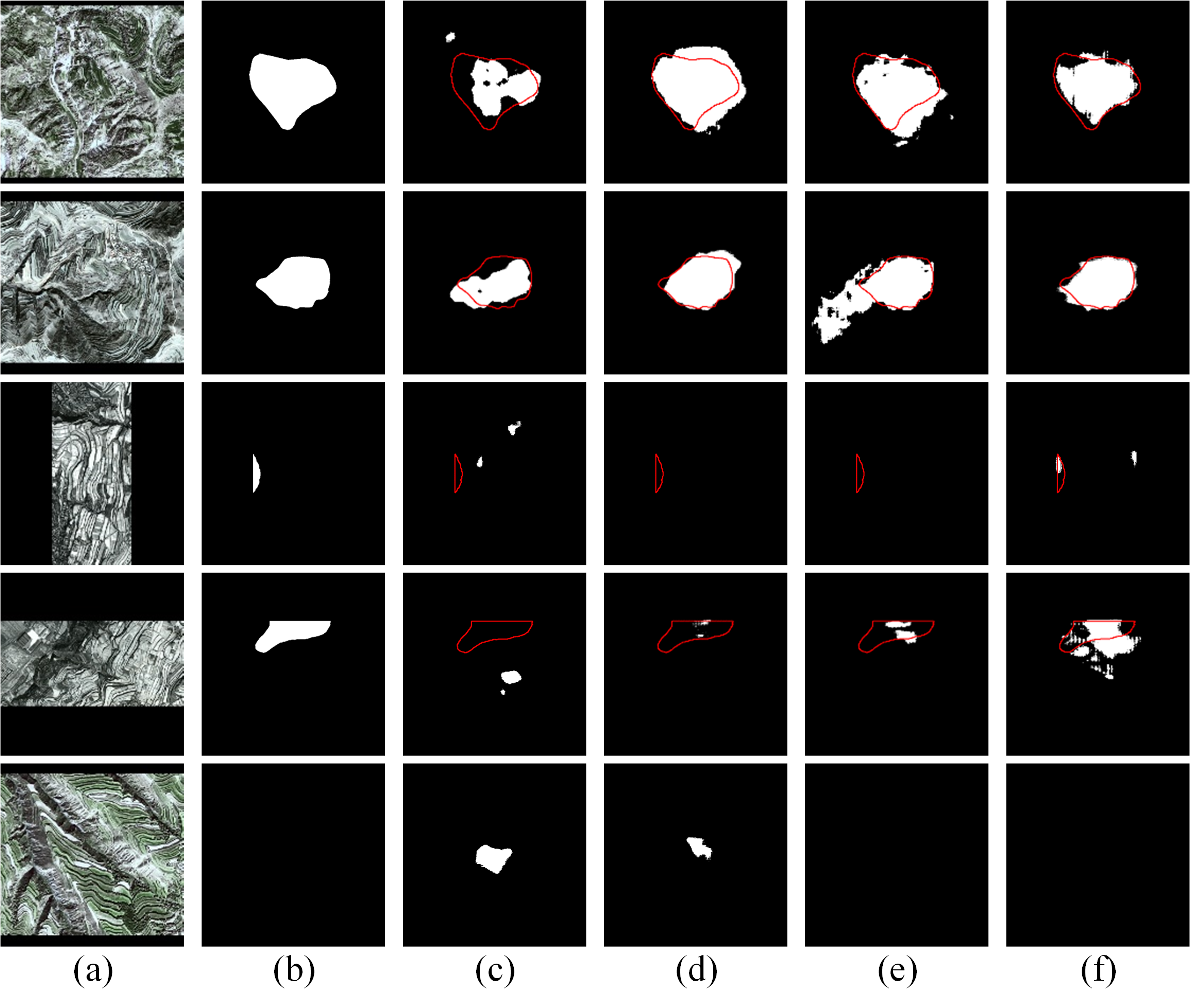}
		\DeclareGraphicsExtensions.
		\caption{Visual results by segmentation cross validation. (a): RGB img; (b): label; (c): baseline; (d): after first round; (e): after second round warm-up; (f): after second round.}
		\label{SegmentationIteration}
	\end{figure}

	The strategies of alternate training of two networks and iterative optimization of the encoder play an important role in the proposed model. We record the performance of the segmentation and classification networks during the model training in Table~\ref{ClassificationIteration} and Fig.~\ref{SegmentationIteration}.
	As can be seen, the second-round performance of both networks is greatly improved compared with the first round, and the third-round performance is similar to the second round. This result is consistent with our design principle and verifies the effectiveness of the iterative training strategy.
	
	Particularly, in the second warm-up stage of the segmentation network, the pixel-level prediction accuracy reaches a high level, demonstrating that the classification network and segmentation networks have achieved unification and sharing of the feature space to a certain extent by using the iterative training strategy.
	
	\subsection{Complexity analysis}
	\begin{table}[h]
		\centering
		\caption{Trainable parameters and complexity of the models.}
		\begin{tabular}{ccc}
			\hline
			& Params (M) & GFLOPs \bigstrut\\
			\hline
			Upper branch & 62.92 & 128.59 \bigstrut[t]\\
			Lower branch & 63.68 & 150.03 \bigstrut[b]\\
			\hline
		\end{tabular}%
		\label{parameters}%
	\end{table}

	We also conduct a complexity analysis of the proposed model in terms of model parameters, GFLOPs, and runtime, where FLOPs stand for a billion floating-point operations per second. The results are listed in Table~\ref{parameters}, which are based on input HRSI pictures of dimension $512 \times 512 \times 3$. Moreover, the total time consumption is shown in Table~\ref{time}.
	
	\begin{table}[h]
		\centering
		\caption{Total time consumption.}
		\begin{tabular}{lc}
			\hline
			\multicolumn{1}{c}{Method} & Traing time (min) \bigstrut\\
			\hline
			Baseline & 62.4 \bigstrut[t]\\
			SOCL  & 192.8 \\
			OCL  & 380.5 \\
			ICSSN  & 674.1 \bigstrut[b]\\
			\hline
		\end{tabular}%
		\label{time}%
	\end{table}
	Compared to the baseline model, the SOCL model consumes almost three times more time because of the extra calculation of the contrast loss function. Due to the introduction of the iterative training strategy, both the OCL network and ICSSN need to be trained several times, which leads to increased more training time. Taking into account the performance improvement, we believe this linear increase in computation is acceptable.
	
	\section{Conclusion}
	This paper proposed the ICSSN for old landslide detection using HRSI data, which is characterized by iterative optimization between the classification newtwork and segmentation network to achieve effective extraction of global abstract information and local detailed information. In the proposed ICSSN, the classification network employs the OCL strategy, uses the double-branch parameter-sharing siamese network as the feature extractor, can simultaneously process a group of inputs, extract and fuse two input features, and realize the learning of the global features of a landslide with the aid of the classifier. In addition, the SOCL module is designed to assist in segmentation network training. This module frames and screens the features of key areas such as the side and back walls of a landslide, and uses the feature extractor the loss auxiliary training for improving the learning of the detailed features of the landslide. A collaborative model training strategy was designed, which involves alternate training of the two networks and iterative optimization of the shared encoder by both networks. The proposed model was evaluated sufficiently on a real-world old landslide dataset via comparative and cross-validation experiments. Extensive experimental results were presented to demonstrate that the proposed ICSSN is able to greatly improve the landslide detection performance at both object and pixel levels.
	

\end{document}